\documentclass[a4paper,fleqn]{cas-dc}

\usepackage[numbers]{natbib}
\usepackage{multicol}
\def\tsc#1{\csdef{#1}{\textsc{\lowercase{#1}}\xspace}}
\tsc{WGM}
\tsc{QE}
\tsc{EP}
\tsc{PMS}
\tsc{BEC}
\tsc{DE}

\begin{document}
\let\WriteBookmarks\relax
\def\floatpagepagefraction{1}
\def\textpagefraction{.001}
\shorttitle{}
\shortauthors{T. Kumar et~al.}

\title [mode = title]{Saliency-Based diversity and fairness  Metric and FaceKeepOriginalAugment: A Novel Approach for Enhancing Fairness and Diversity }                      \tnotemark[1]

\tnotetext[1]{This research was supported by Science Foundation Ireland under grant numbers 18/CRT/6223 (SFI Centre for Research Training in Artificial intelligence), SFI/12/RC/2289/$P\_2$ (Insight SFI Research Centre for Data Analytics), 13/RC/2094/$P\_2$ (Lero SFI Centre for Software) and 13/RC/2106/$P\_2$ (ADAPT SFI Research Centre for AI-Driven Digital Content Technology). For the purpose of Open Access, the author has applied a CC BY public copyright licence to any Author Accepted Manuscript version arising from this submission.}


\author[1]{Teerath Kumar}[]
\cormark[1]
\fnmark[1]
\ead{teerath.menghwar2@mail.dcu.ie}
\ead[url]{https://www.crt-ai.ie/team/teerath-kumar/}
\credit{Data curation, Methdology preparation, Experimental work,   Writing - Original draft preparation}

\affiliation[1]{CRT-AI, 
School of Computing, Dublin City University, 
                city={Dublin},
                country={Ireland}}

\author[2]{Alessandra Mileo}[]
\fnmark[2]
\ead[URL]{https://www.dcu.ie/computing/people/alessandra-mileo}
\credit{Supervised, Reviewed and edited, feedback on method}

\author[3]{Malika Bendechache}[%
   ]
\fnmark[3]
\ead[URL]{https://www.universityofgalway.ie/science-engineering/staff-profiles/malikabendechache/}

\credit{ Supervised, Novel Metric idea, Writing - Original draft preparation, reviewed and edited}

\affiliation[2]{INSIGHT and I-Form Research Centre,
School of Computing,
Dublin City University,
                city={Dublin},
                country={Ireland}}

\affiliation[3]{ADAPT and Lero Research Centres,
School of Computer Science,
University of Galway, 
                city={Galway},
                country={Ireland}}

\cortext[cor1]{Corresponding author}


\begin{abstract}
Data augmentation has become a pivotal tool in enhancing the performance of computer vision tasks, with the KeepOriginalAugment method emerging as a standout technique for its intelligent incorporation of salient regions within less prominent areas, enabling augmentation in both regions. Despite its success in image classification, its potential in addressing biases remains unexplored. In this study, we introduce an extension of the KeepOriginalAugment method, termed FaceKeepOriginalAugment, which explores various debiasing aspects—geographical, gender, and stereotypical biases—in computer vision models. By maintaining a delicate balance between data diversity and information preservation, our approach empowers models to exploit both diverse salient and non-salient regions, thereby fostering increased diversity and debiasing effects. We investigate multiple strategies for determining the placement of the salient region and swapping perspectives to decide which part undergoes augmentation. Leveraging the Image Similarity Score (ISS), we quantify dataset diversity across a range of datasets, including Flickr Faces HQ (FFHQ), WIKI, IMDB, Labelled Faces in the Wild (LFW), UTK Faces, and Diverse Dataset. We evaluate the effectiveness of FaceKeepOriginalAugment in mitigating gender bias across CEO, Engineer, Nurse, and School Teacher datasets, utilizing the Image-Image Association Score (IIAS) in convolutional neural networks (CNNs) and vision transformers (ViTs). Our findings shows the efficacy of FaceKeepOriginalAugment in promoting fairness and inclusivity within computer vision models, demonstrated by reduced gender bias and enhanced overall fairness. Additionally, we introduce a novel metric, Saliency-Based Diversity and Fairness Metric, which quantifies both diversity and fairness while handling data imbalance across various datasets.  
\end{abstract}

\maketitle
\section{Introduction}
Deep learning has shown remarkable success across various domains, such as image processing~\cite{kumarforged, roy2023wildect, roy2022computer, ranjbarzadeh2023me, aleem2022random, kumar2024image}, audio analysis~\cite{park2020search, kumar2020intra, chandio2021audd, turab2022investigating, turab2206investigating, kumar2023audrandaug}, and numerous other fields~\cite{chandio2022precise, baea2021class, raj2023understanding, khan2023sql, khan2022introducing, kumar2024keeporiginalaugment, vavekanand2024cardiacnet, singh2024efficient, singh2023deep, kumar2023advanced, raj2024oxml, kumar2024navigating}. But bias has been in each domain. 
Computer vision models and their applications often exhibit various social biases, including gender bias~\cite{buolamwini2018gender, birhane2021multimodal}, geographical bias~\cite{mandal2021dataset,mandal2023gender}, and racial bias~\cite{buolamwini2018gender,karkkainen2021fairface}. For instance, facial recognition systems tend to be less accurate for individuals with darker skin tones and for females~\cite{buolamwini2018gender}. 
When deploying a model trained on a dataset that doesn't represent the demographics of the patient population, biases can skew results. For instance, if a model trained on mammograms of white patients is used on non-white patients with higher breast density, it may show decreased sensitivity, leading to more missed or delayed diagnoses and potentially worse outcomes for non-White patients~\cite{tejani2024understanding}.

The root cause of these biases often lies in the datasets used to train the models, which propagate biases when deployed in real-time applications. Dataset compilation methods typically involve gathering images from the internet, leading to the creation of biased datasets~\cite{karras2019ffhq}. The auditing of visual datasets for faces has primarily focused on race and gender~\cite{mandal2021dataset,buolamwini2018gender,karkkainen2021fairface}, highlighting the importance of addressing biases in facial regions. 

\begin{figure*}[h!]
    
    \includegraphics[page =1, width=1.1\linewidth,clip, trim=2cm 4.5cm 0cm 4cm]{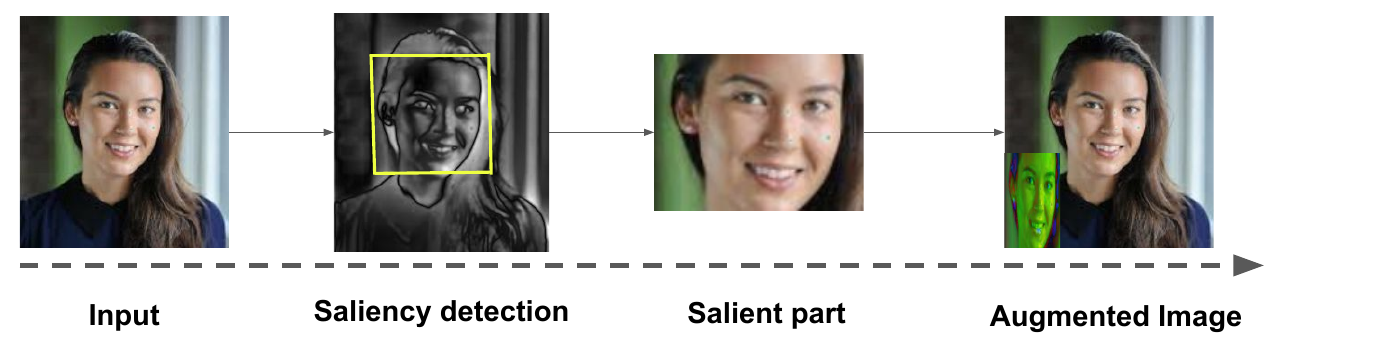}
     \caption{Overall architecture of the proposed approach}
    \label{fig:architecture}
\end{figure*}

\begin{figure*}[hpt!]
    \centering
    \includegraphics[page =2, width=0.8\linewidth, height=8cm]{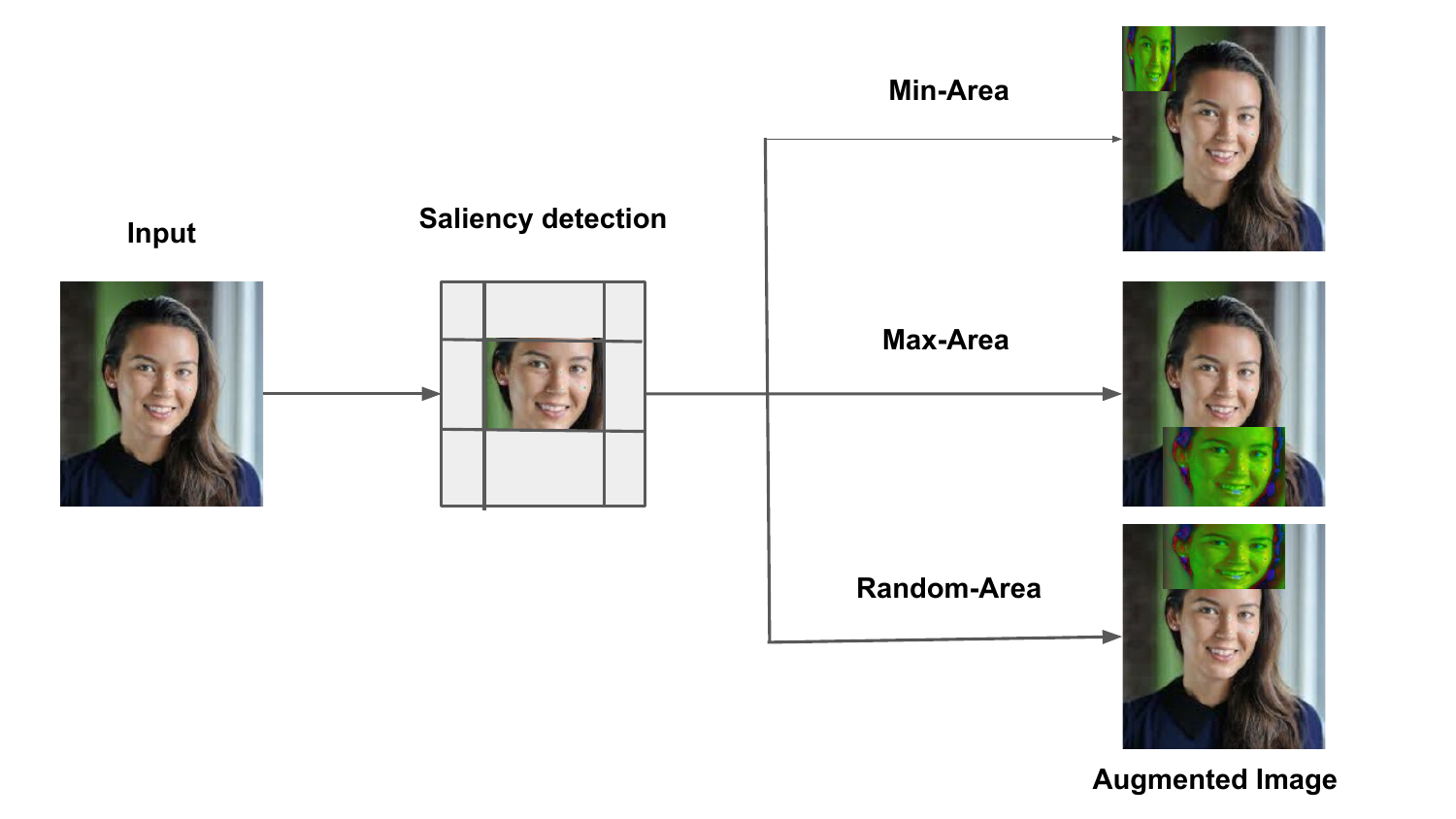}
   \caption{Where to place the salient region?}
    \label{fig:placement}
    \includegraphics[page =3, width=0.8\linewidth, height=8cm]{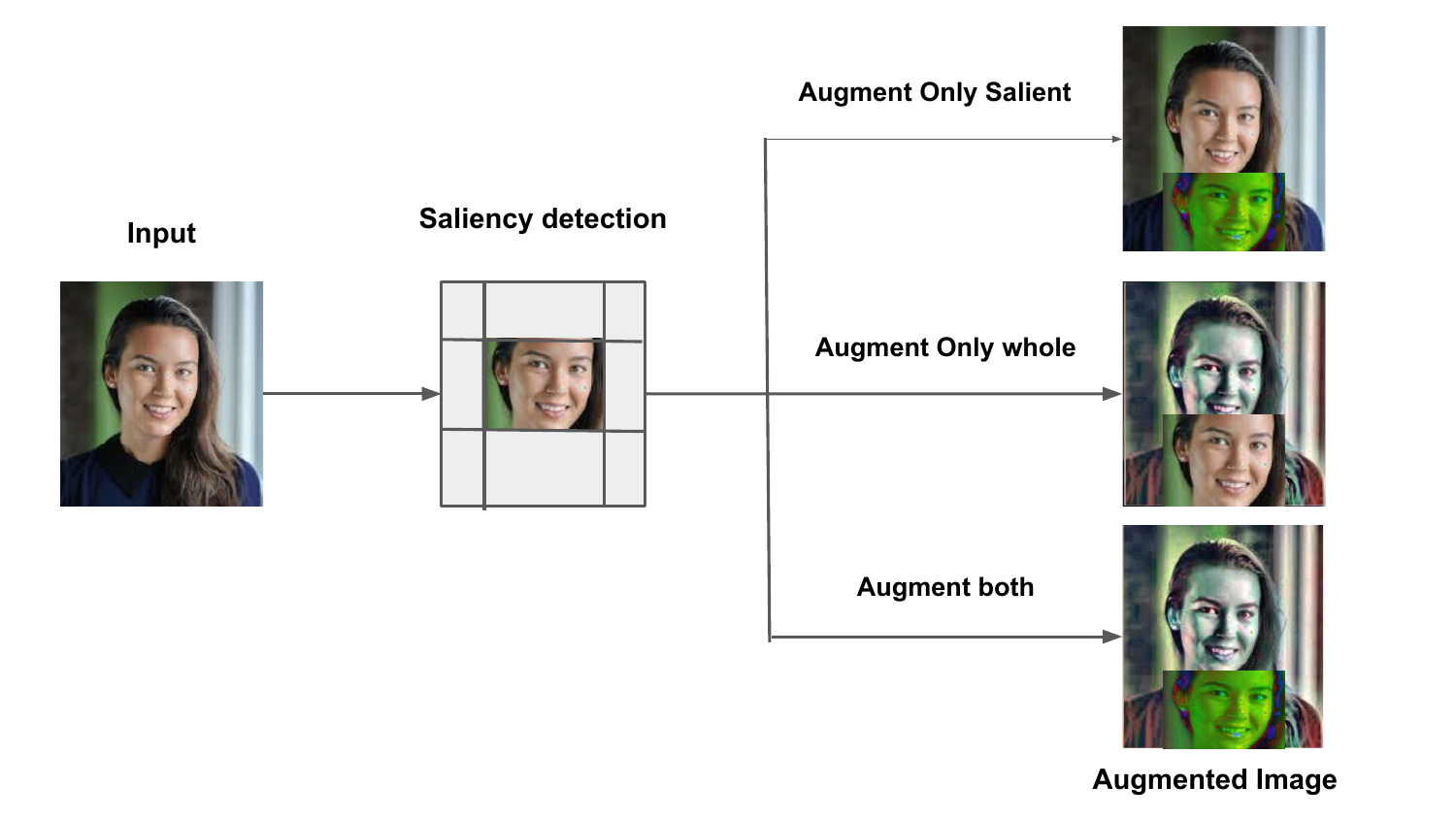}
    \caption{Which part should be augmented?}
    \label{fig:augment_strategy}
\end{figure*}

To mitigate these biases, several methods have been proposed. Zhang et al.~\cite{zhang2020towards} investigated debiasing in image classification tasks by generating adversarial examples to balance training data distribution. Kim et al.~\cite{kim2021biaswap} introduced Biaswap, a method that debiases deep neural networks without prior knowledge of bias types, utilizing unsupervised sorting and style transfer techniques. Lee et al.~\cite{lee2021learning} proposed DiverseBias, a feature-level data augmentation technique for improving the debiasing of image classification models by synthesizing diverse bias-conflicting samples through disentangled representation learning. Other related approaches including SalfMix \cite{choi2021salfmix}, KeepAugment \cite{gong2021keepaugment}, and Randaugment \cite{cubuk2020randaugment}, have also tackled feature fidelity challenges. However, it's worth noting that while these methods enhance data diversity, they may introduce noise and compromise feature diversity, thereby affecting overall model performance. For example, SalfMix may induce overfitting by repetitively incorporating the salient part of the image \cite{choi2021salfmix}, and KeepAugment might introduce a domain shift between salient and non-salient regions, impeding contextual information \cite{gong2021keepaugment}.

In this work, we present a pioneering approach named \textit{FaceKeepOriginalAugment}, detect the salient region and place in any non-salient region as shown in Fig.~\ref{fig:architecture}, aimed at tackling biases in computer vision.  We delve into various strategies for determining the optimal placement of the salient region and meticulously assess the efficacy of our approach in mitigating biases across diverse datasets and occupational categories. In this endeavor, we address the following pivotal questions by extending our previous work~\cite{Kumar2024KeepOriginalAugmentSI}:

\begin{itemize}
  \item To what extent does FaceKeepOriginalAugment enhance dataset diversity, as measured by the Image Similarity Score (ISS), across various datasets like Flickr Faces HQ (FFHQ), WIKI, IMDB, Labelled Faces in the Wild (LFW), and UTK?

\item To what extend does FaceKeepOriginalAugment contribute to mitigate gender bias in CNNs and ViTs, thereby promoting fairness in computer vision models across various datasets?

\item Can we measure diversity and fairness in dataset while dealing data imbalance ?

\end{itemize}

The remainder of our work is organized as follows: Section~\ref{sec:related_work} discusses the relevant literature, Section~\ref{sec:methodology} describes our methodology, Section~\ref{sec:experimental_setup} outlines the experimental setup and results, and finally, Section~\ref{sec:conclusion} concludes our work.

\section{Related work}\label{sec:related_work}

Efforts have been made to address biases in computer vision models. Several studies have shown gender, racial and geographical biases in computer vision tasks. Buolamwini and Gebru \cite{buolamwini2018gender} discovered flaws of facial recognition systems, mainly in detecting faces of dark-skinned women, which implies the importance of bias mitigation in face identification. Mitigation of these challenges has led to several proposals including the use of Fairface \cite{karkkainen2021fairface} which is a curated dataset as well as filtering and debiasing internet-based training datasets \cite{birhane2021multimodal}. Moreover, data augmentation techniques are important for helping alleviate biases by endorsing dataset diversity while maintaining salient information. Information erasing augmentation techniques such as random erasing \cite{zhong2020random}, cutout \cite{devries2017improved}, grid mask \cite{chen2020gridmask}, hide-and-seek \cite{singh2018hide} aim at making models learn erased features by removing given information from images. Nevertheless, these approaches may result into overfitting through noisy examples hence not providing enough diversity. However, there also exist information preserving augmentation methods such as  SalfMix \cite{choi2021salfmix} as well as KeepAugment \cite{gong2021keepaugment} have problems like overfitting and domain shift between non-salient and salient areas, respectively. Our method FaceKeepOriginalAugment builds on the state-of-the-art method KeepOriginalAugment~\cite{Kumar2024KeepOriginalAugmentSI}, which solve the limitations of SalfMix and KeepAugment. Through intelligent inclusion of salient regions into non-salient ones with diverse positioning during augmentation, our approach prevents bias and enhances dataset variation without compromising the truthfulness of features. Although Kumar et al.'s previous work~\cite{Kumar2024KeepOriginalAugmentSI} demonstrated impressive efficacy in classification and related tasks, this study expands upon their approach by investigating its debiasing potential. In addition to examining the debiasing capabilities of the method, we undertake a thorough analysis of the hyperparameters linked to the data augmentation process in debiasing aspect. By integrating debiasing mechanisms into the existing framework and examining the influence of hyperparameters, our objective is to thoroughly explore the debiasing aspect of FaceKeepOriginalAugment through comprehensive experiments conducted on various datasets.

Recent research highlights the critical intersection of diversity and fairness in machine learning, addressing the implications of biased datasets on model performance. Dablain et al.~\cite{dablain2024towards} emphasize the necessity of balancing class instances while mitigating protected feature bias through their Fair OverSampling (FOS) method, which utilizes SMOTE for class imbalance reduction. Zhao et al.~\cite{zhao2024position}  highlighted the importance of establishing clear definitions and rigorous evaluations of the diversity of datasets, focusing on the social constructs inherent in the dataset curation process. Mandal et al.~\cite{mandal2023multimodal} investigated gender bias in multimodal models such as CLIP, employing cosine similarity to measure stereotypical associations in outputs. Their findings reveal the limitations of traditional metrics in capturing the complexity of biases inherent in these models. Similarly, Kuncheva and Whitaker ~\cite{kuncheva2003measures} explored diversity measures within classifier ensembles, illustrating the challenges of correlating diversity with ensemble precision. Together, these works underscore the need for innovative metrics that holistically address both diversity and fairness, advancing the discourse on ethical machine learning practices.

\section{Methodology}\label{sec:methodology}
In this section, we explain our method name \textit{FaceKeepOriginalAugment} and new matric  named \textit{Saliency-Based Diversity and Fairness Metric}. 

\subsection{FaceKeepOriginalAugment}

Our primary objective is to enhance diversity by preserving both the original and augmented information within a single image,  thereby promoting fairness to encourage the model to learn diverse information. To achieve this, we begin by detecting the salient region in image~\footnote{https://docs.opencv.org/3.4/da/dd0/classcv\_1\_1saliency\_1\_1StaticSaliencyFineGrained.html}. In our method, we utilize the saliency detection technique proposed by Montabone et al. \cite{montabone2010human} and Kumar et al. \cite{Kumar2024KeepOriginalAugmentSI}, which has demonstrated superior performance compared to other methods \cite{uddin2020saliencymix}. Our choice of this method is inspired by recent research \cite{uddin2020saliencymix}, which thoroughly investigated different saliency detection methods for data augmentation.  After finding the salient region, we discuss two important questions i) Where to place the salient region?  ii) Which part should be augmented?

 By identifying the important region, we aim to address the issue of redundant salient features present in SalfMix, as well as tackle the domain shift problem encountered in KeepAugment.

To determine the placement of the salient region within the non-salient region and achieve diversity, we explore three different strategies:

\begin{itemize}
    
\item \textbf{Where to place the salient region?}

\textbf{Min-Area:} We identify eight regions surrounding the salient region and select the one with the minimum area. The salient region is resized according to the size of that minimum area and placed within it, as shown in output of first row of Fig.~\ref{fig:placement}.

\textbf{Max-Area:} Conversely, we choose the region with the maximum area among the eight surrounding regions and resize the salient region accordingly, as shown in output of second row of Fig.~\ref{fig:placement}.

\textbf{Random-Area:} We adopt a more flexible approach by randomly selecting one of the eight regions. The salient region is resized based on the size of the chosen region and placed within it as shown in output of third row of Fig.~\ref{fig:placement}.


\item \textbf{Which part should be augmented?}

After identifying the salient region, we propose and investigate three distinct strategies to enhance  diversity:

\textbf{Augment only salient:} We solely apply random augmentation to the salient region and then paste the augmented salient region into the non-salient region of the original image as shown in output of first row of Fig.~\ref{fig:augment_strategy}.

\textbf{Augment non-salient only:} We conduct augmentation on the entire image while preserving the original salient region. The augmented image is then combined with the original salient region, extracted from the unaltered original image as shown in output of second row of Fig.~\ref{fig:augment_strategy}.

\textbf{Augment both:} This strategy involves performing separate augmentations on both the salient region and the entire image. The augmented salient region is integrated with the augmented whole image as shown in output of third row of Fig.~\ref{fig:augment_strategy}.

We observed that the augment both strategy demonstrates greater diversity across various computer vision tasks, detailed discussion is given in section~\ref{hyperparamter}.

It is important to note that we utilize randAug~\cite{cubuk2020randaugment} for augmentation, offering computational efficiency similar to that employed by KeepAugment.
\end{itemize}

\begin{table*}[hpt!]
\centering
\caption{Comparison of Strategies Across Professions for Finding Optimal Hyperparameters - Augmentation Strategy and Area Strategy}
\label{tab:hyperparameters}
\begin{tabular}{l l c c c c c c}
\toprule
\multirow{2}{*}{\textbf{Profession}} & \multirow{2}{*}{\textbf{Strategy}} & \multicolumn{3}{c}{$ISS_{intra}$} & \multicolumn{3}{c}{$ISS_{cross}$} \\ 
\cmidrule(lr){3-5} \cmidrule(lr){6-8} 
                           &             & \textbf{max} & \textbf{min} & \textbf{random} & \textbf{max} & \textbf{min} & \textbf{random} \\ 
\midrule
\multirow{3}{*}{CEO} 
& Only Original & 0.9928 & 0.9935 & 0.9887 & 0.9932 & 0.9959 & 0.9945 \\ 
& Only Salient  & 0.9933 & 0.9913 & 0.9915 & 0.9960 & 0.9935 & 0.9926 \\ 
& Both          & 0.9901 & 0.9928 & \textbf{1.0701} & 0.9946 & 0.9949 & \textbf{1.0809} \\ 
\midrule
\multirow{3}{*}{Nurse} 
& Only Original & 0.9944 & 0.9947 & 0.9943 & 0.9969 & 0.9953 & 0.9974 \\ 
& Only Salient  & 0.9959 & 0.9955 & 0.9932 & 0.9948 & 0.9945 & 0.9947 \\ 
& Both          & 0.9943 & 0.9934 & \textbf{1.0521} & 0.9943 & 0.9944 & \textbf{1.0057} \\ 
\midrule
\multirow{3}{*}{Engineer} 
& Only Original & 0.9960 & 0.9970 & 0.9999 & 0.9982 & 0.9974 & 0.9965 \\ 
& Only Salient  & 1.0000 & 0.9921 & 0.9999 & 0.9958 & 0.9972 & 0.9965 \\ 
& Both          & 0.9964 & 0.9971 & \textbf{1.0170} & 0.9978 & 0.9963 & \textbf{1.0304} \\ 
\midrule
\multirow{3}{*}{School Teacher} 
& Only Original & 0.9963 & 0.9934 & 0.9968 & 0.9971 & 0.9968 & 0.9968 \\ 
& Only Salient  & 0.9946 & 0.9974 & 0.9967 & 0.9972 & 0.9961 & 0.9981 \\ 
& Both          & 0.9948 & 0.9969 & \textbf{1.0558} & 0.9962 & 0.9977 & \textbf{1.0552} \\ 
\midrule
\multirow{3}{*}{Politician} 
& Only Original & 0.9952 & 0.9963 & 0.9927 & 0.9981 & 0.9952 & 0.9942 \\ 
& Only Salient  & 0.9943 & 0.9940 & 0.9930 & 0.9963 & 0.9947 & 0.9962 \\ 
& Both          & 0.9939 & 0.9943 & \textbf{1.0832} & 0.9967 & 0.9959 & \textbf{1.0353} \\ 
\bottomrule
\end{tabular}
\end{table*}

\begin{table}[hpt!]
\centering
\caption{ISS$_{intra}$ of Datasets and baseline result originate from~\cite{mandal2021dataset}.}
\label{tab:iss_intra}
\begin{tabular}{l c c}
\toprule
\textbf{Dataset} & \textbf{Baseline} & \textbf{Ours} \\
\midrule
FFHQ~\cite{karras2019ffhq} & $0.9940$ & \textbf{0.9951} \\
Diverse Dataset~\cite{mandal2021dataset} & $0.9895$ & \textbf{0.9965} \\
WIKI~\cite{rothe2015dex} & $0.9786$ & \textbf{0.9980} \\
IMDB~\cite{rothe2015dex} & $0.9661$ & \textbf{0.9971} \\
LFW~\cite{learned2016labeled} & $0.9536$ & \textbf{0.9956} \\
UTK~\cite{zhang2017age} & $0.9418$ & \textbf{0.9903} \\
\bottomrule
\end{tabular}
\end{table}

\subsection{Saliency-Based Diversity and Fairness Metric}

This work introduces a novel Saliency-Based Diversity and Fairness Metric designed to account for both within-group diversity and inter-group fairness. Within-group diversity measures the variation among samples within a single group, capturing the richness of diversity in each class. Inter-group fairness evaluates the differences between groups, ensuring equitable representation and mitigating bias. The proposed metric integrates both aspects, weighting them appropriately to handle imbalanced datasets while maintaining fairness and diversity across groups. Mandal et al.\cite{mandal2021dataset} used cosine similarity, which measures the angle between two vectors, focusing on the direction rather than the magnitude. It works well for comparing the similarity between two feature vectors in terms of orientation, but for data diversity, we generally need to capture more than just the direction.

We refer to the feature vectors as \(X\). First, we perform saliency detection on the images and then pass these saliency image to the pretrained VGG16 model~\cite{simonyan2014very}, created by the Visual Geometry Group at the University of Oxford, to obtain the feature vectors. Before calculating the diversity metrics, the feature vectors are normalized to ensure consistency and comparability across groups. The normalization of the feature vectors is performed as follows:
\begin{equation}
X' = \frac{X}{\|X\|}    
\end{equation}

where $X'$ is the normalized feature vector and $\|X\|$ is its Euclidean norm. By normalizing the feature vectors, we ensure that the diversity metrics are scale-invariant and comparable across different groups. The normalized feature vectors are then used to calculate Euclidean distances. This ensures that both within-group and inter-group diversity measures are normalized and lie within a comparable range.
\subsubsection{Within-Group Diversity}

Let $X'_i$ represent the set of normalized saliency-based features for group $i$ (e.g., Male or Female). The within-group diversity for group $i$ is computed using the Euclidean distance between pairs of feature vectors within that group:

\begin{equation}
D_{\text{within}}(X'_i) = \frac{1}{N_i(N_i - 1)} \sum_{j=1}^{N_i} \sum_{k=j+1}^{N_i} \text{dist}(X'_{i,j}, X'_{i,k})
\end{equation}

where $\text{dist}(X'_{i,j}, X'_{i,k})$ is the Euclidean distance between two normalized feature vectors $X'_{i,j}$ and $X'_{i,k}$, and $N_i$ is the number of feature vectors in group $i$.

\subsubsection{Inter-Group Diversity and Fairness}

The inter-group diversity measures the average pairwise distances between feature vectors from different groups (e.g., between Male and Female groups). This aspect is crucial for ensuring that the metric captures inter-group fairness. Let $X'_i$ and $X'_j$ represent the feature sets of two different groups. The inter-group diversity is computed as:

\begin{equation} 
D_{\text{inter}}(X'_i, X'_j) = \frac{1}{N_i N_j} \sum_{k=1}^{N_i} \sum_{l=1}^{N_j} \text{dist}(X'_{i,k}, X'_{j,l})
\end{equation}

where $N_i$ and $N_j$ are the number of feature vectors in groups $i$ and $j$, respectively. The Euclidean distance between normalized feature vectors ensures that the inter-group diversity reflects the actual distance between different groups. 

\subsubsection{Combined Metric: Fairness and Diversity}
In scenarios where the dataset is imbalanced (i.e., the number of samples across groups varies significantly), it is important to ensure that larger groups do not dominate the overall diversity metric. To handle this, the metric weights the within-group and inter-group diversity terms by the size of each group, allowing the metric to reflect the actual contributions of both minority and majority groups.

The final metric, which combines both diversity within groups and diversity across groups, is given in equation.~\ref{eq:fairness_diversity}. The feature normalization and the use of group sizes in weighting ensure that the final metric is balanced and reflects the contributions of both smaller and larger groups while also keeping the metric bounded between 0 and 1 (assuming both $\alpha$ and $\beta$ are less than 0.5 individually). 

\twocolumn[ 
\begin{center}
\begin{equation}
M_{\text{fairness-diversity}} = \alpha \cdot \frac{1}{N} \sum_{i=1}^{K} N_i \cdot D_{\text{within}}(X'_i) + \beta \cdot \frac{1}{N(N-1)} \sum_{i=1}^{K} \sum_{j=i+1}^{K} N_i \cdot N_j \cdot D_{\text{inter}}(X'_i, X'_j)
\label{eq:fairness_diversity}
\end{equation}
\end{center}
]

where:
- $K$ is the total number of groups,
- $N_i$ is the number of samples in group $i$,
- $N = \sum_{i=1}^{K} N_i$ is the total number of samples across all groups,
- $\alpha$ is the weight for within-group diversity (focusing on intra-group diversity),
- $\beta$ is the weight for inter-group diversity (focusing on fairness across groups).

\begin{table*}[hpt!]
\centering
\caption{Image Similarity score across all possible queries. Baseline results are taken from~\cite{mandal2021dataset}.}
\label{tab:various_queries}
\begin{tabular}{l l c c}
\toprule
\multirow{2}{*}{\textbf{Query}} & \multirow{2}{*}{\textbf{Language Location Pair}} & \multicolumn{2}{c}{\textbf{ISS$_{Intra}$}} \\
\cmidrule{3-4}
 & & \textbf{Baseline} & \textbf{Ours} \\
\midrule
\multirow{9}{*}{CEO} & Arabic-West Asia \& North Africa & 0.899012 & \textbf{0.9988} \\
 & English-North America & 0.968974 & \textbf{0.9985} \\
 & English-West Europe & 0.929469 & \textbf{0.9964} \\
 & Hindi-South Asia & \textbf{0.997845} & 0.9898 \\
 & Indonesian-SE Asia & \textbf{0.983675} & 0.9912 \\
 & Mandarin-East Asia & \textbf{0.989452} & 0.9927 \\
 & Russian-East Europe & 0.959661 & \textbf{0.9957} \\
 & Spanish-Latin America & 0.974743 & \textbf{0.9929} \\
 & Swahili-Sub Saharan Africa & 0.977119 & \textbf{0.9936} \\
\midrule
\multirow{9}{*}{Engineer} & Arabic-West Asia \& North Africa & 0.98639 & \textbf{0.9959} \\
 & English-North America & 0.988344 & \textbf{1.0031} \\
 & English-West Europe & 1.000911 & 1.0004 \\
 & Hindi-South Asia & \textbf{1.003149} & 0.9966 \\
 & Indonesian-SE Asia & \textbf{0.987191} & 0.9897 \\
 & Mandarin-East Asia & \textbf{0.991146} & 0.9923 \\
 & Russian-East Europe & \textbf{1.007155} & 0.9976 \\
 & Spanish-Latin America & 0.984955 & \textbf{0.9992} \\
 & Swahili-Sub Saharan Africa & 0.983727 & \textbf{0.9957} \\
\midrule
\multirow{9}{*}{Nurse} & Arabic-West Asia \& North Africa & \textbf{1.002607} & 0.9916 \\
 & English-North America & 0.971564 & \textbf{0.9933} \\
 & English-West Europe & 0.99561 & \textbf{0.9982} \\
 & Hindi-South Asia & 0.984535 & \textbf{0.9968} \\
 & Indonesian-SE Asia & 0.975914 & \textbf{0.9937} \\
 & Mandarin-East Asia & 0.98904 & \textbf{0.9983} \\
 & Russian-East Europe & \textbf{0.997979} & 0.9940 \\
 & Spanish-Latin America & \textbf{1.000587} & 0.9964 \\
 & Swahili-Sub Saharan Africa & 0.958532 & \textbf{0.9950} \\
\midrule
\multirow{9}{*}{Politician} & Arabic-West Asia \& North Africa & 0.977348 & \textbf{0.9951} \\
 & English-North America & 0.995927 & \textbf{0.9987} \\
 & English-West Europe & 0.979358 & \textbf{0.9968} \\
 & Hindi-South Asia & 0.979915 & \textbf{0.9936} \\
 & Indonesian-SE Asia & 0.972307 & \textbf{0.9921} \\
 & Mandarin-East Asia & 0.976251 & \textbf{0.9957} \\
 & Russian-East Europe & 0.93835 & \textbf{0.9992} \\
 & Spanish-Latin America & 0.988452 & \textbf{0.9943} \\
 & Swahili-Sub Saharan Africa & 0.943626 & \textbf{0.9983} \\
\midrule
\multirow{9}{*}{\begin{tabular}{l} School \\ Teacher \end{tabular}} & Arabic-West Asia \& North Africa & 1.014298 & 1.0000 \\
 & English-North America & 0.997715 & 0.9978 \\
 & English-West Europe & 0.940142 & \textbf{0.9994} \\
 & Hindi-South Asia & 1.000047 & 0.9972 \\
 & Indonesian-SE Asia & 0.985991 & \textbf{1.0017} \\
 & Mandarin-East Asia & 1.00862 & 0.9984 \\
 & Russian-East Europe & 0.976169 & \textbf{0.9952} \\
 & Spanish-Latin America & 0.965902 & \textbf{0.9980} \\
 & Swahili-Sub Saharan Africa & 0.985919 & \textbf{1.0033} \\
\bottomrule
\end{tabular}
\end{table*}

\begin{table}[hpt!]
\vspace{-15pt}
\centering
\caption{Image Similarity score across all possible queries. Baseline results are taken from~\cite{mandal2021dataset}.}
\label{tab:iss_scores}
\begin{tabular}{l c c c c}
\toprule
& \multicolumn{2}{c}{ISS$_{intra}$} & \multicolumn{2}{c}{ISS$_{cross}$} \\
\cmidrule(lr){2-3} \cmidrule(lr){4-5}
\multicolumn{1}{c}{Query} & Baseline & Ours & Baseline & Ours \\
\midrule
CEO & 0.9644 & \textbf{0.9944} & 0.9846 & \textbf{0.9956} \\
Engineer & 0.9925 & \textbf{0.9967} & 0.9939 & \textbf{0.9972} \\
Nurse & 0.9862 & \textbf{0.9952} & 0.9900 & \textbf{0.9961} \\
Politician & 0.9724 & \textbf{0.9960} & 0.9836 & \textbf{0.9964} \\
School Teacher & 0.9860 & \textbf{0.9990} & 0.9904 & \textbf{0.9977} \\
\midrule
Mean Value & 0.9803 & \textbf{0.9963} & 0.9885 & \textbf{0.9966} \\
\bottomrule
\end{tabular}
\vspace{-1pt} 
\end{table}

\begin{table*}[hpt!]
\centering
\caption{Average Image-Image Association Scores (IIAS) for CNNs and ViTs. Positive values indicate bias towards men, negative towards women. Total absolute IIAS reflects bias magnitude. Our approach reduces gender bias, highlighted in red, baseline results are taken from~\cite{mandal2023biased}}
\label{tab:gender_bias}
\begin{tabular}{l c c c c c c c c}
\toprule
\multicolumn{9}{c}{\textbf{Masked}} \\
\midrule
\multirow{2}{*}{Class} & \multicolumn{4}{c}{Biased} & \multicolumn{4}{c}{Unbiased} \\
\cmidrule(lr){2-9}
& $\underset{\text{Baseline}}{\mathrm{CNN}}$ & $\underset{\text{Ours}}{\mathrm{CNN}}$ & $\underset{\text{Baseline}}{\mathrm{ViT}}$ & $\underset{\text{Ours}}{\mathrm{ViT}}$ & $\underset{\text{Baseline}}{\mathrm{CNN}}$ & $\underset{\text{Ours}}{\mathrm{CNN}}$ & $\underset{\text{Baseline}}{\mathrm{ViT}}$ & $\underset{\text{Ours}}{\mathrm{ViT}}$ \\
\midrule
CEO & 0.059 & \textbf{0.025} & 0.1 & \textbf{0.027} & 0.26 & \textbf{0.037} & 0.02 & \textbf{0.022} \\
Engineer & 0.23 & \textbf{0.023} & 0.14 & \textbf{0.006} & 0.36 & \textbf{0.043} & 0.17 & \textbf{0.018} \\
Nurse & -0.14 & \textbf{-0.040} & -0.35 & \textbf{-0.011} & -0.05 & \textbf{-0.053} & -0.2 & \textbf{-0.023} \\
School Teacher & -0.17 & \textbf{-0.038} & -0.15 & \textbf{-0.006} & -0.12 & \textbf{-0.077} & -0.05 & \textbf{-0.04} \\
\midrule
\multirow{2}{*}{Total IIAS (abs)} & 0.599 & \textbf{0.126} & 0.74 & \textbf{0.05} & 0.79 & \textbf{0.21} & 0.44 & \textbf{0.103} \\
\cmidrule(lr){2-9}
Bias reduction & \multicolumn{2}{c}{\textcolor{red}{$\sim 5$ times $\downarrow$}} & \multicolumn{2}{c}{\textcolor{red}{$\sim 15$ times $\downarrow$}} & \multicolumn{2}{c}{\textcolor{red}{$\sim 4$ times $\downarrow$}} & \multicolumn{2}{c}{\textcolor{red}{$\sim 4$ times $\downarrow$}} \\
\midrule
\multicolumn{9}{c}{\textbf{Unmasked}} \\
\midrule
\multirow{2}{*}{Class} & \multicolumn{4}{c}{Biased} & \multicolumn{4}{c}{Unbiased} \\
\cmidrule(lr){2-9}
& $\underset{\text{Baseline}}{\mathrm{CNN}}$ & $\underset{\text{Ours}}{\mathrm{CNN}}$ & $\underset{\text{Baseline}}{\mathrm{ViT}}$ & $\underset{\text{Ours}}{\mathrm{ViT}}$ & $\underset{\text{Baseline}}{\mathrm{CNN}}$ & $\underset{\text{Ours}}{\mathrm{CNN}}$ & $\underset{\text{Baseline}}{\mathrm{ViT}}$ & $\underset{\text{Ours}}{\mathrm{ViT}}$ \\
\midrule
CEO & 0.050 & \textbf{0.023} & 0.17 & \textbf{0.003} & 0.07 & \textbf{0.03} & 0.023 & \textbf{0.059} \\
Engineer & 0.180 & \textbf{0.016} & 0.19 & \textbf{0.008} & 0.04 & \textbf{0.036} & 0.21 & \textbf{0.003} \\
Nurse & -0.21 & \textbf{-0.039} & -0.21 & \textbf{-0.035} & -0.06 & \textbf{-0.023} & -0.17 & \textbf{-0.003} \\
School Teacher & -0.02 & \textbf{-0.035} & -0.4 & \textbf{-0.001} & -0.04 & \textbf{-0.056} & -0.14 & \textbf{-0.036} \\
\midrule
\multirow{2}{*}{Total IIAS (abs)} & 0.46 & \textbf{0.113} & 0.97 & \textbf{0.047} & 0.21 & \textbf{0.145} & 0.58 & \textbf{0.101} \\
\cmidrule(lr){2-9}
Bias reduction & \multicolumn{2}{c}{\textcolor{red}{$\sim 4$ times $\downarrow$}} & \multicolumn{2}{c}{\textcolor{red}{$\sim 21$ times $\downarrow$}} & \multicolumn{2}{c}{\textcolor{red}{$\sim 2$ times $\downarrow$}} & \multicolumn{2}{c}{\textcolor{red}{$\sim 6$ times $\downarrow$}} \\
\bottomrule
\end{tabular}
\end{table*}

\section{Experiments}\label{sec:experimental_setup}

\subsection{Experimental setup}
To assess data diversity in terms of geographical and stereotypical biases, we employ two variants of the Image Similarity Score (ISS): ISS$_{Intra}$ and ISS$_{Cross}$. ISS$_{Intra}$ quantifies data diversity within a dataset, while ISS$_{Cross}$ evaluates diversity across different datasets, as introduced by Mandal et al.~\cite{mandal2021dataset}. Range of both ISS metrics is 0 to 2.  We use data diversity term as more data diversity lead to less bias~\cite{mandal2021dataset}. We utilize five datasets, namely, Flickr Faces HQ (FFHQ), WIKI, IMDB, Labelled Faces in the Wild (LFW), UTK Faces, and Diverse Dataset, maintaining the same experimental settings as detailed in \cite{mandal2021dataset}.

To measure gender bias in both Convolutional Neural Networks (CNNs) and Vision Transformers (ViTs), we employ the Image-Image Association Score (IIAS) \cite{mandal2023multimodal,mandal2023biased}. IIAS evaluates bias by comparing the similarity between gender attributes in images using cosine similarity and images representing specific concepts, as outlined in \cite{mandal2023multimodal}. Our experimental setup encompasses four CNN models (VGG16, ResNet152, Inceptionv3, and Xception) and four ViT models (ViT B/16, B/32, L/16, and L/32). We adhere to the training settings outlined in \cite{mandal2023gender}, wherein for CNNs, layers are initially frozen and custom layers are trained for 50 epochs.  For ViTs, layers are first frozen and trained for 100 epochs, followed by training all layers for 50 epochs with a low learning rate.

A gender bias study, as detailed in \cite{mandal2023gender}, collected images from Google searches using job terms, resulting in two training sets: one with equal gender representation and another with biased representation. The test set remains balanced. The training dataset comprises 7,200 images (900 per category), while the test dataset consists of 1,200 images (300 per category, 150 per gender). Additionally, separate datasets for men and women were utilized for evaluation purposes. We integrate our proposed FaceKeepOriginalAugment approach into the transformations used in the experiments, addressing biases.

\subsection{Hyperparamter}\label{hyperparamter}
We conducted an extensive analysis of various hyperparameter combinations for five professional datasets—CEO, Engineer, Nurse, Politician, and School Teacher—using two distinct data diversity metrics: ISS$_{Intra}$ and ISS$_{Cross}$. Specifically, we investigated the combinations of augmentation strategies with area strategies. The ISS$_{Intra}$ and ISS$_{cross}$ scores for CEO, Engineer, Nurse, Politician, and School Teacher are shown in Table~\ref{tab:hyperparameters}. Our analysis revealed that the augmentation \textit{both strategy} with the \textit{random area strategy} yielded the optimal scores in both metrics, consistent with findings reported in \cite{Kumar2024KeepOriginalAugmentSI}. The reasons could be, the selection of the random area strategy is attributed to its provision of scaling augmentation, while the augment both strategy enhances diversity, making it particularly effective for debiasing.

\begin{table*}[hpt!]
\centering
\caption{Summary of Diversity and Fairness Metrics for Balanced and Imbalanced Datasets}
\label{tab:diversity_dataset_fairness_metrics}
\begin{tabular}{lccc|ccc}
\toprule
Dataset & 
\multicolumn{3}{c|}{Balanced} & \multicolumn{3}{c}{Imbalanced} \\
\cmidrule(lr){2-4} \cmidrule(lr){5-7}
 & $D_{within}$ & $D_{inter}$ & $M_{\text{fairness-diversity}}$ & $D_{within}$ & $D_{inter}$ & $M_{\text{fairness-diversity}}$ \\
 \midrule
\multicolumn{7}{c}{\textbf{Baseline}} \\ 
\midrule
DiverseDataset & 0.83$\pm$ 0.01 & 0.35$\pm$ 0.00 & 0.59$\pm$ 0.00 & 0.82$\pm$ 0.02 & 0.35$\pm$ 0.00 & 0.59$\pm$ 0.01 \\
FFHQ & 0.83$\pm$ 0.02 & 0.35$\pm$ 0.00 & 0.59$\pm$ 0.01 & 0.82$\pm$ 0.01 & 0.32$\pm$ 0.00 & 0.57$\pm$ 0.01 \\
IMDB & 0.80$\pm$ 0.01 & 0.35$\pm$ 0.00 & 0.57$\pm$ 0.00 & 0.80$\pm$ 0.01 & 0.34$\pm$ 0.00 & 0.57$\pm$ 0.00 \\
LFW & 0.84$\pm$ 0.02 & 0.36$\pm$ 0.00 & 0.60$\pm$ 0.01 & 0.81$\pm$ 0.01 & 0.27$\pm$ 0.00 & 0.54$\pm$ 0.01 \\
UTK & 0.80$\pm$ 0.01 & 0.35$\pm$ 0.00 & 0.57$\pm$ 0.00 & 0.80$\pm$ 0.01 & 0.35$\pm$ 0.00 & 0.57$\pm$ 0.00 \\
WIKI & 0.84$\pm$ 0.01 & 0.36$\pm$ 0.00 & 0.60$\pm$ 0.00 & 0.80$\pm$ 0.01 & 0.23$\pm$ 0.00 & 0.52$\pm$ 0.00 \\
\midrule

\multicolumn{7}{c}{\textbf{With FaceKeepOriginalAugment}} \\ 
\midrule
DiverseDataset & 0.83$\pm$ 0.00 & 0.36$\pm$ 0.00 & 0.59$\pm$ 0.00 & 0.82$\pm$ 0.01 & 0.35$\pm$ 0.00 & 0.59$\pm$ 0.00 \\
FFHQ & 0.83$\pm$ 0.02 & 0.36$\pm$ 0.00 & 0.59$\pm$ 0.01 & 0.83$\pm$ 0.01 & 0.32$\pm$ 0.00 & 0.57$\pm$ 0.00 \\
IMDB & 0.81$\pm$ 0.01 & 0.35$\pm$ 0.00 & 0.58$\pm$ 0.00 & 0.81$\pm$ 0.01 & 0.34$\pm$ 0.00 & 0.57$\pm$ 0.00 \\
LFW & 0.83$\pm$ 0.01 & 0.36$\pm$ 0.00 & 0.59$\pm$ 0.01 & 0.81$\pm$ 0.01 & 0.27$\pm$ 0.00 & 0.54$\pm$ 0.00 \\
UTK & 0.81$\pm$ 0.02 & 0.35$\pm$ 0.00 & 0.58$\pm$ 0.01 & 0.80$\pm$ 0.01 & 0.35$\pm$ 0.00 & 0.58$\pm$ 0.00 \\
WIKI & 0.84$\pm$ 0.01 & 0.36$\pm$ 0.00 & 0.60$\pm$ 0.00 & 0.82$\pm$ 0.00 & 0.24$\pm$ 0.00 & 0.53$\pm$ 0.00 \\

\bottomrule
\end{tabular}
\end{table*}

\begin{table*}[hpt!]
\centering
\caption{Diversity and Fairness Metrics measurement of different Language Location pair queries across gender }
\label{tab:diversity_fairness_metrics}
\begin{tabular}{lccc|ccc}
\toprule
\textbf{Language Location Pairs} & 
\multicolumn{3}{c|}{\textbf{Baseline}} & \multicolumn{3}{c}{\textbf{With FaceKeepOriginalAugmentaiton}} \\
\cmidrule(lr){2-4} \cmidrule(lr){5-7}
 & $D_{within}$ & $D_{inter}$ & $M_{\text{fairness-diversity}}$ & $D_{within}$ & $D_{inter}$ & $M_{\text{fairness-diversity}}$ \\
 \midrule  
Arabic-West Asia \& North Africa & 0.59$\pm$ 0.27 & 0.49$\pm$ 0.16 & 0.54$\pm$ 0.06 & 0.79$\pm$ 0.01 & 0.35$\pm$ 0.00 & 0.57$\pm$ 0.01 \\
English - North America & 0.59$\pm$ 0.27 & 0.49$\pm$ 0.16 & 0.54$\pm$ 0.06 & 0.81$\pm$ 0.01 & 0.35$\pm$ 0.00 & 0.58$\pm$ 0.01 \\
English-West Europe & 0.59$\pm$ 0.28 & 0.50$\pm$ 0.17 & 0.55$\pm$ 0.06 & 0.80$\pm$ 0.02 & 0.35$\pm$ 0.00 & 0.57$\pm$ 0.01 \\
Hindi-South Asia & 0.59$\pm$ 0.27 & 0.48$\pm$ 0.15 & 0.53$\pm$ 0.06 & 0.80$\pm$ 0.01 & 0.35$\pm$ 0.00 & 0.57$\pm$ 0.00 \\
Indonesian-South East Asia & 0.59$\pm$ 0.27 & 0.49$\pm$ 0.16 & 0.54$\pm$ 0.06 & 0.79$\pm$ 0.01 & 0.35$\pm$ 0.00 & 0.57$\pm$ 0.01 \\
Mandarin-East Asia & 0.60$\pm$ 0.27 & 0.51$\pm$ 0.18 & 0.55$\pm$ 0.04 & 0.79$\pm$ 0.01 & 0.35$\pm$ 0.00 & 0.57$\pm$ 0.00 \\
Russian-East Europe & 0.59$\pm$ 0.27 & 0.50$\pm$ 0.17 & 0.54$\pm$ 0.05 & 0.80$\pm$ 0.01 & 0.35$\pm$ 0.00 & 0.57$\pm$ 0.01 \\
Spanish-Latin America & 0.60$\pm$ 0.27 & 0.49$\pm$ 0.16 & 0.54$\pm$ 0.06 & 0.79$\pm$ 0.00 & 0.35$\pm$ 0.00 & 0.57$\pm$ 0.00 \\
Swahili-Sub Saharan Africa & 0.59$\pm$ 0.27 & 0.50$\pm$ 0.17 & 0.55$\pm$ 0.05 & 0.80$\pm$ 0.01 & 0.35$\pm$ 0.00 & 0.57$\pm$ 0.00 \\

\bottomrule

\end{tabular}
\end{table*}

\begin{table*}[hpt!]
\centering
\caption{Diversity and Fairness Metrics measurement of different Profession datasets across Language Location pairs}
\label{tab:diversity_fairness_professions}
\begin{tabular}{lccc|ccc}
\toprule
\textbf{Profession} & 
\multicolumn{3}{c|}{\textbf{Baseline}} & \multicolumn{3}{c}{\textbf{With FaceKeepOriginalAugmentaiton}} \\
\cmidrule(lr){2-4} \cmidrule(lr){5-7}
 & $D_{within}$ & $D_{inter}$ & $M_{\text{fairness-diversity}}$ & $D_{within}$ & $D_{inter}$ & $M_{\text{fairness-diversity}}$ \\
 \midrule
CEO & 0.83$\pm$ 0.01 & 0.61$\pm$ 0.00 & 0.72$\pm$ 0.00 & 0.86$\pm$ 0.00 & 0.63$\pm$ 0.00 & 0.74$\pm$ 0.00 \\
Engineer & 0.83$\pm$ 0.01 & 0.62$\pm$ 0.00 & 0.73$\pm$ 0.00 & 0.85$\pm$ 0.00 & 0.63$\pm$ 0.00 & 0.74$\pm$ 0.00 \\
Nurse & 0.82$\pm$ 0.01 & 0.61$\pm$ 0.00 & 0.72$\pm$ 0.01 & 0.86$\pm$ 0.00 & 0.63$\pm$ 0.00 & 0.74$\pm$ 0.00 \\
Politician & 0.82$\pm$ 0.00 & 0.61$\pm$ 0.00 & 0.72$\pm$ 0.00 & 0.86$\pm$ 0.00 & 0.62$\pm$ 0.00 & 0.74$\pm$ 0.00 \\
School Teacher & 0.83$\pm$ 0.00 & 0.62$\pm$ 0.00 & 0.72$\pm$ 0.00 & 0.86$\pm$ 0.01 & 0.62$\pm$ 0.00 & 0.74$\pm$ 0.00 \\

\bottomrule
\end{tabular}
\end{table*}

\subsection{Results}

To assess dataset diversity, we employ the Intra-dataset Image Similarity Score (ISS$_{intra}$) across various datasets, showcasing substantial improvements compared to the baseline as shown in Table \ref{tab:iss_intra}. To further anaylse, the Table~\ref{tab:various_queries} presents the ISS$_{intra}$ for various queries across different language-location pairs. The baseline ISS values are compared with ours. The queries include CEO, Engineer, Nurse, Politician, and School Teacher, each with multiple language-location pairs. It's observed that the ISS values for most queries and language-location pairs have improved compared to the baseline values. This improvement indicates the effectiveness of the proposed method  in enhancing image similarity in cross-cultural contexts. Notably, certain queries show more significant improvements, such as CEO and School Teacher in Arabic-West Asia \& North Africa, Mandarin-East Asia, and Spanish-Latin America language locations as shown in Table~\ref{tab:various_queries}.
Furthermore, we investigate the Inter-dataset Image Similarity Score (ISS$_{cross}$) and ISS$_{intra}$ for five occupation datasets. Our approach consistently demonstrates enhanced diversity, with notable exceptions observed in the school teacher dataset, potentially influenced by underlying biases identified by Mandal et al. (2023) \cite{mandal2023gender}. Specifically, our method outperforms the baseline across different datasets, notably achieving significant improvements for occupations such as "Politician" and "Nurse". While both approaches exhibit comparable performance for "Politician" in ISS$_{cross}$, our method showcases superior results across diverse occupations, emphasizing its effectiveness in promoting dataset diversity as shown in Table~\ref{tab:iss_scores}. Moreover, our approach yields higher mean ISS scores across all queries, highlighting its efficacy in enhancing diversity. Overall, our approach presents promising advancements in dataset diversity assessment as shown in Table~\ref{tab:iss_scores}.

\begin{figure*}[hpt!]
    \centering
    \includegraphics[width=0.9\textwidth, height=11cm, trim= 3cm 0cm 0cm 0cm]{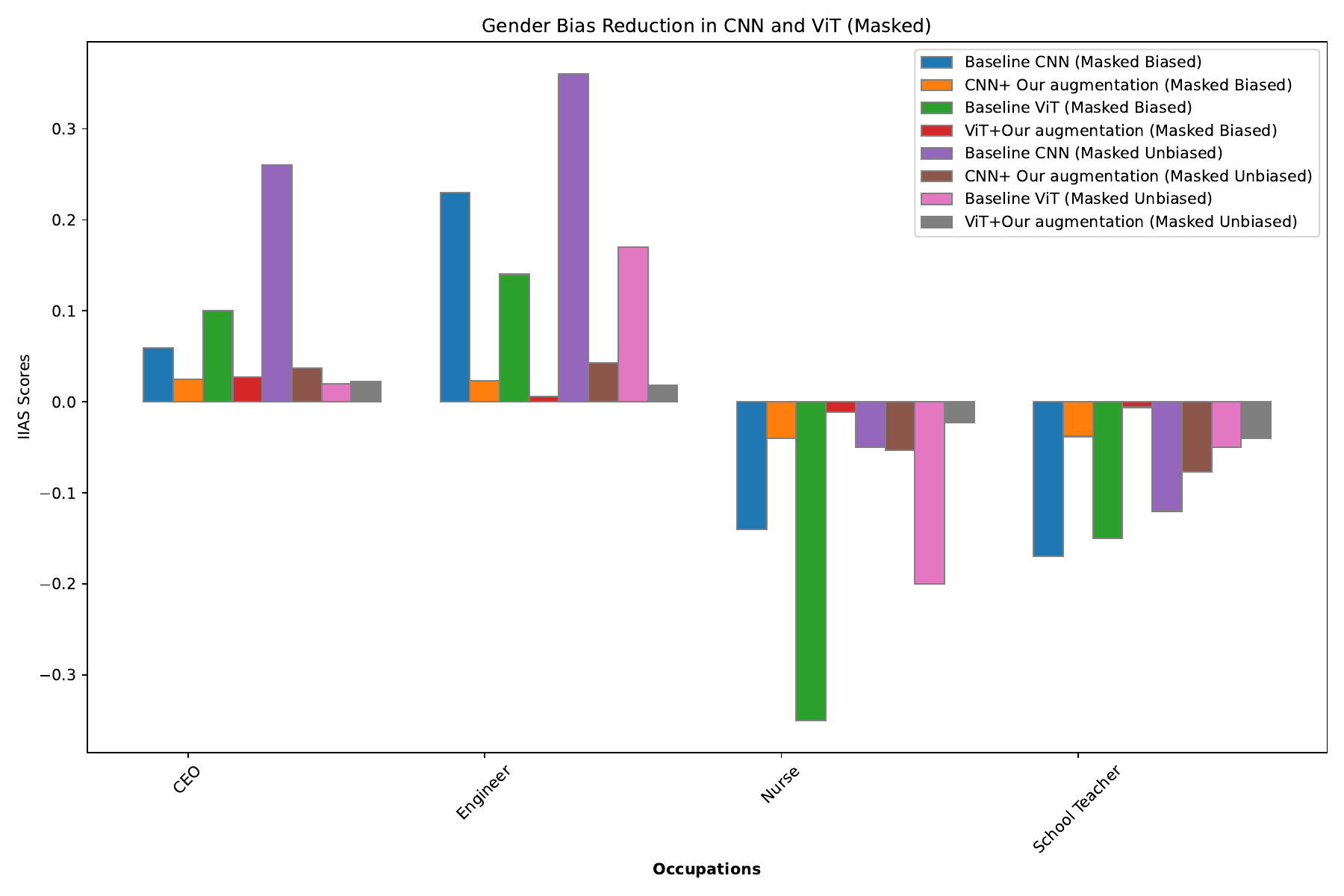}
    \caption{Comparison of our approach for gender bias reduction in CNN and ViT- Masked Scienario. }
    \label{fig:masked_biased}

    \centering
    \includegraphics[width=0.9\textwidth, height=11cm, trim= 3cm 0cm 0cm 0cm]{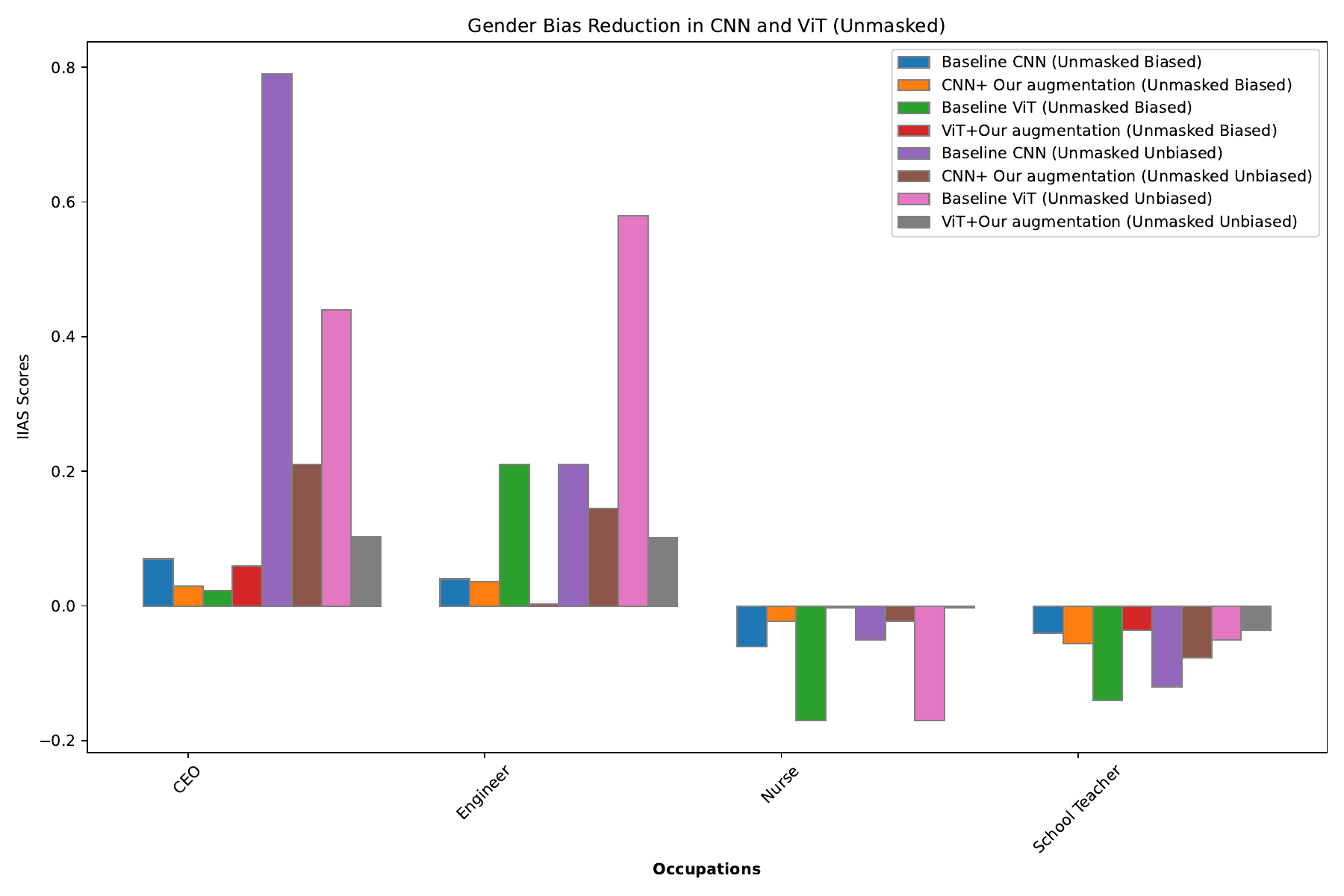}
    \caption{Comparison of our approach for gender bias reduction in CNN and ViT- Unmasked Scienario. }
    \label{fig:unmasked_biased}
\end{figure*}

\begin{figure*}[hpt!]
    \centering
    \includegraphics[width=1.0\linewidth,trim= 0cm 0cm 0cm 2.0cm,clip]{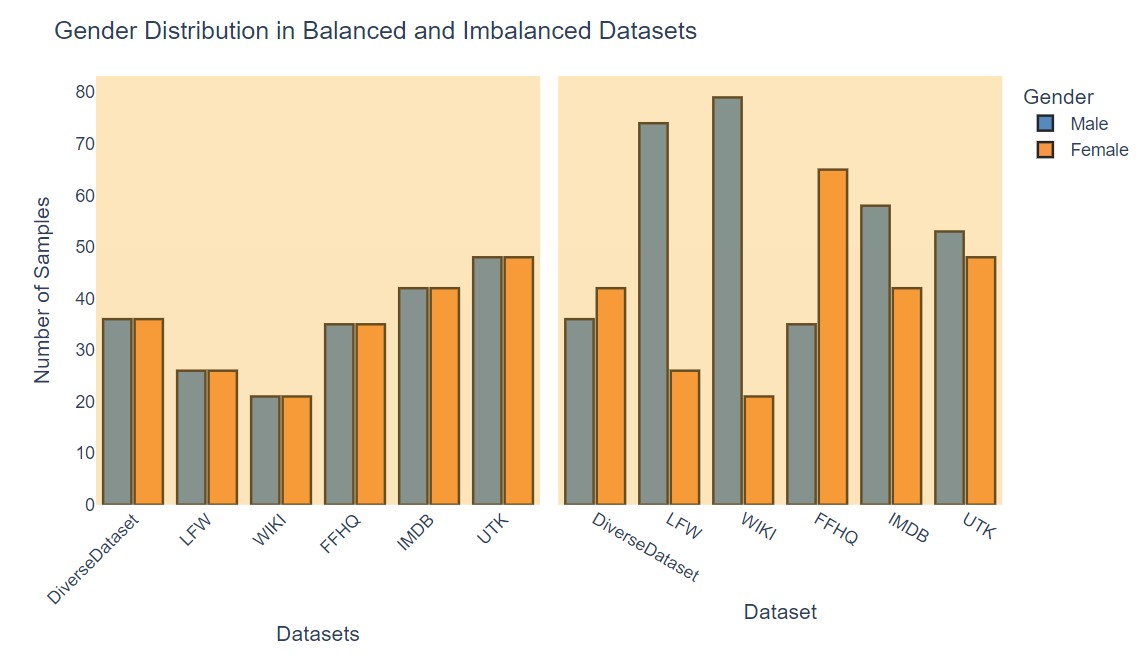}
    \caption{Number of the samples in Balance and Imbalance datasets }
    \label{fig:balance_imblanace}
\end{figure*}
Table~\ref{tab:gender_bias} presents the average Image-Image Association Scores (IIAS) for Convolutional Neural Networks (CNNs) and Vision Transformers (ViTs) across various classes and masking scenarios. Positive IIAS values indicate biases towards men, while negative values signify biases towards women. Notably, in the biased dataset, biases towards men are observed for the CEO and Engineer classes, as evidenced by positive IIAS values, while biases towards women are observed for the Nurse and School Teacher classes, indicated by negative IIAS values. This nuanced analysis reveals the gender biases inherent in the dataset and underscores the importance of addressing these biases in computer vision models. FaceKeepOriginalAugment approach consistently demonstrates a remarkable reduction in gender bias compared to baseline models. Specifically, our method achieves reductions of approximately 5 to 15 times for CNNs and ViTs in masked scenarios as shown in figure~\ref{fig:masked_biased}, and 4 to 21 times in unmasked scenarios as shown in figure~\ref{fig:unmasked_biased} across various occupation classes. These substantial reductions, highlighted in red, underscore the effectiveness of our approach in promoting fairness and inclusivity within computer vision models. Moreover, it's crucial to note the total absolute IIAS values, which reflect the overall magnitude of gender bias within the models. Our approach consistently yields lower total absolute IIAS values compared to baseline results taken from~\cite{mandal2023biased}, indicating a substantial reduction in the overall magnitude of gender bias. This comprehensive view of bias reduction further reinforces the robustness and efficacy of FaceKeepOriginalAugment in mitigating gender bias within computer vision models. Our findings not only highlight the nuanced gender biases present across different occupation classes but also demonstrate the significant effectiveness of our approach in addressing these biases. By reducing gender bias and promoting fairness within computer vision models, our FaceKeepOriginalAugment approach contributes towards building more bias-free system.

\subsection{Saliency-Based Diversity and Fairness Metric Evaluation}

To evaluate the proposed Saliency-Based Diversity and Fairness Metric, we conducted experiments on six datasets commonly used for assessing gender diversity in images. These datasets include the Diverse Dataset~\cite{mandal2021dataset}, FFHQ~\cite{karras2019ffhq}, WIKI~\cite{rothe2015dex}, IMDB~\cite{rothe2015dex}, LFW~\cite{learned2016labeled}, and UTK~\cite{zhang2017age}. For the experiments, we utilized the full set of 81 images from the Diverse Dataset and randomly selected 100 images from each of the remaining five datasets. The images were manually classified into two groups: male and female. To address the inherent class imbalance in these datasets, we created balanced versions using an undersampling data augmentation technique, as illustrated in Fig.~\ref{fig:balance_imblanace}.

In addition to gender-based diversity, we extended our evaluation to measure diversity and fairness across nine geographically distinct language-location datasets~\cite{mandal2023multimodal,mandal2021dataset}, which include: West Asia \& North Africa, North America, Western Europe, South Asia, South East Asia, East Asia, Eastern Europe, Latin America, and Sub-Saharan Africa. For each of these regions, we manually organized the images into two gender groups: male and female.

To further generalize the applicability of the metric beyond gender, we assessed saliency-based diversity and fairness across five professions (CEO, Engineer, Nurse, Politician, and School Teacher) in the same nine language-location combinations as discussed in~\cite{mandal2023multimodal,mandal2021dataset}. Each profession was evaluated across these nine sub-categories to ensure a comprehensive understanding of diversity and fairness in various professional and cultural contexts.

In Table \ref{tab:diversity_dataset_fairness_metrics}, the comparison of balanced and imbalanced datasets across the three key metrics within group diversity ($D_{within}$), between group diversity ($D_{inter}$), and the proposed fairness-diversity metric ($M_{\text{fairness-diversity}}$) highlights the effect of data augmentation strategies. Overall, FaceKeepOriginalAugment demonstrates a slight improvement in $D_{within}$ and $D_{inter}$ across most datasets when compared to the baseline. For instance, in the IMDB dataset, FaceKeepOriginalAugment increases $D_{within}$ from 0.80 to 0.81 and maintains $D_{inter}$ at 0.35 for balanced datasets. Similarly, in the WIKI dataset, FaceKeepOriginalAugment slightly improves $M_{\text{fairness-diversity}}$ from 0.52 to 0.53 in imbalanced conditions, suggesting improved fairness when dealing with underrepresented groups. The results for FFHQ and LFW datasets remain stable with minimal fluctuations, showcasing that the augmentation method preserves diversity and fairness without significant deviations from the baseline metrics. The notable consistency in $D_{within}$ across diverse datasets implies that KeepOriginalAugment maintains intra-group diversity effectively, a strength of the approach. Meanwhile, $M_{\text{fairness-diversity}}$ is relatively stable across both balanced and imbalanced datasets, indicating robustness to bias. FaceKeepOriginalAugment marginally improves fairness while preserving diversity, offering a balanced solution that enhances representation, especially in datasets where class imbalance poses a challenge. 

In Table \ref{tab:diversity_fairness_metrics}, we present the diversity and fairness metrics, within group diversity ($D_{within}$), between-group diversity ($D_{inter}$), and the fairness-diversity metric ($M_{\text{fairness-diversity}}$) for various language-location pair queries both in their baseline state and with the  FaceKeepOriginalAugment. The baseline metrics exhibit consistent values across different pairs, with $D_{within}$ ranging from 0.59 to 0.60 and $M_{\text{fairness-diversity}}$ values hovering around 0.53 to 0.55, raising fairness concerns  in these datasets. With FaceKeepOriginalAugment, there is a noticeable enhancement in the metrics. For instance, the $D_{within}$ values improve to 0.79 for most pairs, showcasing a substantial increase in intra-group diversity, while $M_{\text{fairness-diversity}}$ also sees an uplift to around 0.57 across the board. Notably, pairs such as {Arabic-West Asia \& North Africa} and English - North America show an increase in $D_{within}$ from 0.59 to 0.79 and $M_{\text{fairness-diversity}}$ from 0.54 to 0.57, respectively. This improvement signifies that the augmentation strategy enhances both diversity and fairness in these pairs, addressing potential biases that might be present in the baseline datasets. The results illustrate that FaceKeepOriginalAugment effectively enriches the diversity while maintaining fairness across gender in various language-location pairs. This highlights the method's strength in fostering a more equitable representation in datasets that may initially show limited diversity. The insights drawn from these metrics suggest that employing such augmentation strategies is crucial in developing models that prioritize fairness and diversity, particularly in linguistically diverse contexts.

In Table \ref{tab:diversity_fairness_professions}, we present the diversity and fairness metrics within group diversity for various professions both in their baseline state and with FaceKeepOriginalAugment.  The baseline results indicate a high level of within-group diversity for each profession, with $D_{within}$ values consistently around 0.82 to 0.83. This suggests that the profession datasets are already fairly diverse in terms of representation. The $M_{\text{fairness-diversity}}$ values for the baseline remain stable at approximately 0.72, indicating a reasonable balance between fairness and diversity across these professions. with FaceKeepOriginalAugment, there is an observable enhancement in the diversity metrics. The $D_{within}$ values increase to between 0.85 and 0.86, reflecting an improvement in intra-group diversity. Similarly, the $M_{\text{fairness-diversity}}$ remains consistent at 0.74 for all professions, demonstrating that the augmentation method maintains fairness while enhancing diversity. Notably, the $D_{inter}$ values show a slight increase in some cases, such as the CEO and Nurse professions, where the values are maintained around 0.61 to 0.63 after augmentation. This suggests that the method does not negatively impact the between-group diversity while improving within-group diversity. Overall, the results indicate that FaceKeepOriginalAugment effectively increases the diversity of profession datasets while preserving fairness across language location pairs. The enhancements in the metrics highlight the importance of using augmentation strategies to ensure a more equitable representation in datasets used for model training, particularly in contexts where professions may be underrepresented or biased.

\section{Conclusion}\label{sec:conclusion}
We present FaceKeepOriginalAugment, an innovative approach designed to mitigate geographical, gender, and stereotypical biases prevalent in computer vision models. Our method adeptly integrates salient regions within non-salient areas, allowing for flexible augmentation across both regions. By striking a balance between data diversity and preservation, FaceKeepOriginalAugment not only enhances dataset diversity but also effectively reduces biases. We conducted rigorous experiments on a range of diverse datasets, including FFHQ, WIKI, IMDB, LFW, UTK Faces, and a custom Diverse Dataset, employing the Intra-Set Similarity (ISS) metric to assess dataset diversity quantitatively. Additionally, we evaluated the effectiveness of FaceKeepOriginalAugment in addressing gender bias across various professions, such as CEO, Engineer, Nurse, and School Teacher, utilizing the Image-Image Association Score (IIAS) metric for a comprehensive assessment. Our results demonstrate that FaceKeepOriginalAugment significantly reduces gender bias in both Convolutional Neural Networks (CNNs) and Vision Transformers (ViTs).

Moreover, we propose a novel saliency-based diversity and fairness metric that allows for a nuanced evaluation of gender, language-location, and profession biases across diverse datasets. By comparing outcomes with and without the application of FaceKeepOriginalAugment, our findings reveal a substantial enhancement in fairness and inclusivity, representing a pivotal advancement in the ongoing efforts to address biases within computer vision applications.

\printcredits

\bibliographystyle{cas-model2-names}

\bibliography{cas-refs}

\vskip3pt

\bio{}
Mr Teerath Kumar received his Bachelor degree in Computer Science with distinction from National University of Computer and Emerging Science (NUCES), Islamabad, Pakistan and received master in Computer Science and Engineering from Kyung Hee University, South Korea. Currently, he is pursuing PhD from Dublin City University, Ireland. His research interests include advanced data augmentation, biasness and fairness in computer vision, deep learning for medical imaging, generative adversarial networks semi-supervised learning and Neuro-Symbolic AI
\endbio

\bio{}
Dr. Alessandra Mileo  is an Associate Professor in the School of Computing at Dublin City University, a Principal Investigator at  the INSIGHT Centre for Data Analytics and a Funded Investigator at the I-Form Centre for Advanced Manufacturing. Alessandra holds a PhD in Computer Science from the University of Milan, Italy. She secured over 1 million euros in national, international and industry-funded projects, publishing 90+ papers in the area of Internet of Things, Knowledge Graphs, Stream Reasoning, Neuro-symbolic computing and Explainable AI

\endbio

\bio{}
Dr Malika Bendechache is a Lecturer/Assist
Professor in the School of Computer Science at the
University of Galway, Ireland, and a Funded Investigator at ADAPT Centre for AI–Driven Digital Content Technology. Malika holds a PhD in Computer Science from University College Dublin, Ireland. Her research interests span the areas of Big data Analytics, Machine Learning, AI Governance \& Data Governance, Security and Privacy. She designs novel Big Data Analytics and Machine Learning techniques to enhance the capability and efficiency of complex systems, and she also leverages complex systems to improve the effectiveness, privacy and trustworthiness of Analytics/Machine Learning techniques.
\endbio

\end{document}